\documentclass[conference]{IEEEtran}
\usepackage{times}

\usepackage[numbers]{natbib}
\usepackage{multicol}
\usepackage[bookmarks=true]{hyperref}

\usepackage{graphicx}
\usepackage{amssymb}
\usepackage{amsmath}
\usepackage[ruled]{algorithm2e}
\usepackage{comment}
\usepackage{xcolor}
\usepackage{booktabs}
\usepackage{subcaption}
\usepackage[table]{xcolor}

\definecolor{pastelblue}{rgb}{0.68, 0.78, 0.81}

\SetCommentSty{mycommfont}

\begin{document}

% paper title
\title{
Variance-Reduced
Model Predictive Path Integral \\
via Quadratic Model Approximation
}
% You will get a Paper-ID when submitting a pdf file to the conference system
% \author{Author Names Omitted for Anonymous Review. Paper-ID 361}

% \begin{comment}
\author{\authorblockN{Fabian Schramm\textsuperscript{1},
Franki Nguimatsia Tiofack\textsuperscript{1},
Nicolas Perrin-Gilbert\textsuperscript{2},
Marc Toussaint\textsuperscript{3} and
Justin Carpentier\textsuperscript{1}}
\vspace{1.5ex}
\authorblockA{\textsuperscript{1}Inria and DI-ENS, PSL Research University \quad
\textsuperscript{2}Sorbonne University \quad
\textsuperscript{3}TU Berlin\\
Email: fabian.schramm@inria.fr
}}
% \end{comment
% }

\maketitle

\begin{abstract}
Sampling-based controllers, such as Model Predictive Path Integral (MPPI) methods, offer substantial flexibility but often suffer from high variance and low sample efficiency. To address these challenges, we introduce a hybrid variance-reduced MPPI framework that integrates a prior model into the sampling process. Our key insight is to decompose the objective function into a known approximate model and a residual term. Since the residual captures only the discrepancy between the model and the objective, it typically exhibits a smaller magnitude and lower variance than the original objective.
Although this principle applies to general modeling choices, we demonstrate that adopting a quadratic approximation enables the derivation of a closed-form, model-guided prior that effectively concentrates samples in informative regions. 
Crucially, the framework is agnostic to the source of geometric information, allowing the quadratic model to be constructed from exact derivatives, structural approximations (e.g., Gauss- or Quasi-Newton), or gradient-free randomized smoothing.
We validate the approach on standard optimization benchmarks, a nonlinear, underactuated cart-pole control task, and a contact-rich manipulation problem with non-smooth dynamics.
Across these domains, we achieve faster convergence and superior performance in low-sample regimes compared to standard MPPI. These results suggest that the method can make sample-based control strategies more practical in scenarios where obtaining samples is expensive or limited.
\end{abstract}

\IEEEpeerreviewmaketitle

\section{Introduction}
\label{sec:introduction}

Formulating control, estimation, and motion planning as optimization problems has become a standard paradigm in automatic control and robotics. 
Traditionally, these problems are posed in their primal form: one seeks a single optimal configuration or trajectory $x^*$ that minimizes a cost function $f(x)$. 
Standard approaches to solving this problem, such as gradient descent or Newton's method, rely on local first- or second-order information. While efficient, these methods assume $f$ is smooth and can converge to local minima when $f$ is non-convex.
To overcome these limitations, an alternative viewpoint elevates the problem to the \emph{space of probability measures}. Here, the objective is to minimize the expected cost under a distribution $\mu$. This measure-theoretic formulation unifies several powerful frameworks in global optimization, stochastic control, and sampling-based motion planning~\cite{Theodorou2012RelativeEA,KAZIM2024100931}. \\

Within this distributional approach, two complementary families of methods have emerged. The first relies on convex relaxations of the infinite-dimensional measure optimization problem. Lasserre’s Moment–Sum-of-Squares (SOS) hierarchy~\cite{lasserre2001}, and its algebraic counterpart by Parrilo~\cite{parrilo2003semidefinite}, provides a systematic sequence of semidefinite programming (SDP) relaxations whose solutions converge to the global optimum under mild regularity assumptions. These methods offer strong theoretical guarantees but remain limited in practice as the size of the SDPs grows rapidly with the dimensionality and polynomial degree of the underlying problem.
Recent extensions, such as KernelSOS \cite{rudi_finding_2024,groudiev2025samplingbasedglobaloptimalcontrol}, address these limitations by replacing polynomial bases with reproducing kernels, enabling the treatment of non-polynomial objectives and infinite-dimensional function spaces. While such approaches significantly broaden the expressive power of SOS relaxations, they still require solving large-scale SDPs, which makes real-time use in high-frequency control loops impractical.
\begin{figure}[t]
    \centering
    \includegraphics[width=0.8\linewidth]{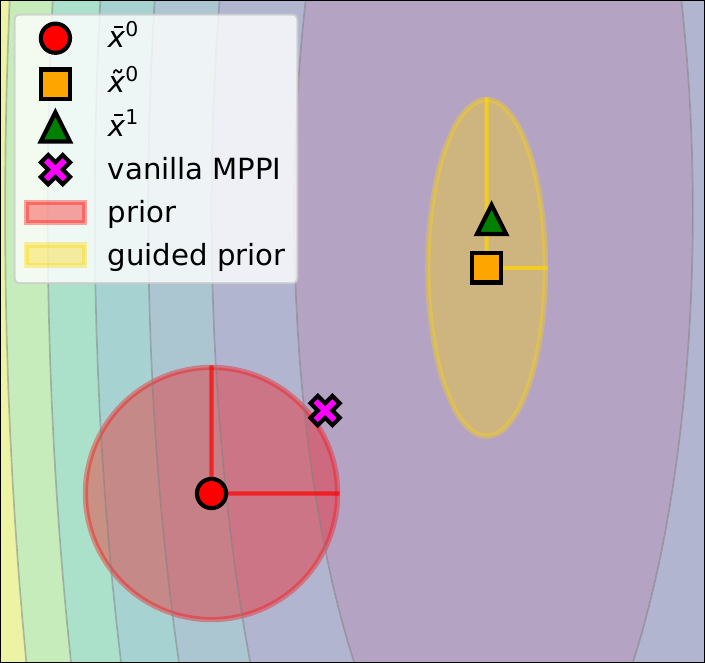}
    \caption{\textbf{Illustration of covariance adaptation via Newton-like approximation.} 
    Model-guided MPPI and vanilla MPPI start at the same state (circle) and use $100$ samples from an isotropic Gaussian prior. 
    The cross indicates the vanilla MPPI update.
    Our method exploits the gradient and Hessian at $\bar{x}^0$ to construct a quadratic model.
    This leads to a guided prior centered at $\tilde{x}^0$ (rectangle), concentrating the sampling distribution (yellow) along the valley and enabling an update $\bar{x}^1$ (triangle) by sampling the residual
    toward the optimum.
    }
    \vspace{-1.5em}
    \label{fig:quadratic-narrow-valley}
\end{figure}

The second family originates from information-theoretic or KL-regularized formulations of stochastic optimal control. Here, the optimal control distribution is expressed as a Boltzmann distribution over trajectories, obtained by introducing an entropy or KL penalty on the control~\cite{Kappen2004LinearTF, Todorov2006LinearlysolvableMD, Theodorou2012RelativeEA}. This reformulation naturally leads to sampling-based algorithms, most notably the Model Predictive Path Integral (MPPI)~\cite{Williams2017ModelPP, Williams2017InformationTheoreticMP} control algorithm. 
MPPI offers compelling practical advantages: it is derivative-free, accommodates non-smooth costs (e.g., contact events), and parallelizes efficiently on modern hardware accelerators. However, unlike convex relaxation methods, MPPI approximates the optimal distribution via Monte Carlo sampling, which introduces a central challenge: high estimator variance. Standard isotropic sampling fails to adapt to landscape geometry, resulting in poor sample efficiency and slow convergence in high-dimensional spaces. 
While recent work has shown that incorporating curvature information by reconstructing local Jacobians and performing Gauss–Newton updates can partially mitigate these effects~\cite{homburger2026gaussnewtonacceleratedmppicontrol}, such approaches remain in a high-sample regime.
Consequently, state-of-the-art implementations often rely on thousands of parallel rollouts to achieve stable performance.

Variance reduction is a well-studied topic in broader stochastic optimization. CMA-ES~\cite{Hansen2001cma, Hansen2006} and the Cross-Entropy Method~\cite{Rubinstein1999TheCM} adapt covariance matrices based on elite samples, while classical techniques like importance sampling~\cite{Rubinstein1981SimulationAT}, control variates~\cite{Glasserman2003MonteCM}, or baseline subtraction in reinforcement learning~\cite{Williams2004SimpleSG,peter2006policygradient}, explicitly target variance reduction in Monte-Carlo estimators. 
Similar ideas appear in large-scale stochastic optimization, where variance-reduced gradient methods and curvature-aware updates are known to accelerate convergence~\cite{Bottou2016OptimizationMF}.

Across domains, incorporating structural information consistently improves the efficiency of noisy estimators. This suggests that similar ideas may also benefit sampling-based control. Yet, unlike many large-scale optimization settings, real-time control imposes strict computational budgets, forcing algorithms to operate in a low-sample regime. These observations raise an important question: \emph{Can we retain the flexibility of sampling-based control while achieving faster convergence with significantly fewer samples?} \\

In this work, we introduce a hybrid approach that guides sampling using a local model of the objective function, thereby accelerating the overall convergence of MPPI-based methods.
Our main contributions are as follows:
(i)~we derive a variance-reduced MPPI framework by decomposing the objective into a known model and a residual term, 
(ii)~we propose a specific instantiation using a quadratic model approximation, deriving a closed-form, model-guided proposal distribution, 
(iii)~we show that this quadratic model can be computed via either exact derivatives, structural approximations, or a gradient-free randomized smoothing scheme, and 
(iv)~we validate the approach numerically, showing superior convergence in low-sample regimes compared to standard baselines.

The remainder of this paper is organized as follows. Section~\ref{sec:preliminaries} formalizes the optimization problem and the standard MPPI derivation. Section~\ref{sec:method} details the derivation of the quadratic model approximation. Section~\ref{sec:experiments} presents the numerical results, and Section~\ref{sec:conclusion} concludes the paper.

\section{Preliminaries}
\label{sec:preliminaries}

\noindent We consider optimization problems of the form:
\begin{equation}
\label{eqn:point-min}
    x^* \in \underset{x \in \Omega}{\arg \min} \, f(x),
\end{equation}
where $f : \Omega \rightarrow \mathbb{R}$ denotes the objective and $\Omega \subseteq \mathbb{R^d}$ represents the feasible set.

From a measure-theoretic perspective, the problem~\eqref{eqn:point-min} can be reformulated as an equivalent problem over measures~\cite{lasserre2001}. 
The goal is to find a probability distribution $\mu^*$, minimizing the expected loss over distributions whose support lies in $\Omega$:
\begin{equation}
\label{eqn:distribution-min}
    \mu^* \in \underset{\substack{\mu \\ \text{supp}(\mu) \subseteq \Omega}}{\arg \min} \, \mathcal{J}(\mu) \quad \text{with} \quad \mathcal{J}(\mu) = \int_{\Omega} f(x)\mu(x) \, dx.
\end{equation}
Depending on the setup, the objective function $f$ can be convex, smooth, or arbitrarily complicated (non-convex, non-smooth). 
For instance, if $f$ is strictly convex with a unique global minimum, working on \eqref{eqn:distribution-min} will converge to the optimal solution, which in the limit is a Dirac \mbox{$\mu^*(x) \to \delta(x - x^*)$} concentrating the distribution mass entirely on the global optimal solution $x^*$. 

While Problem~\eqref{eqn:distribution-min} theoretically allows for finding global optima~\cite{lasserre2001}, solving it over the infinite-dimensional space of measures is generally intractable.
Instead of a direct global solution, a common iterative approach is to perform sequential updates that penalize changes to the current distribution.
At each step $k$, we seek a new distribution $\mu^{k+1}$ that minimizes the original objective $\mathcal{J}(\mu)$ in~\eqref{eqn:distribution-min} subject to a proximity penalization to stay close to the previous estimate $\mu^k$:
\begin{equation}
    \mu^{k+1} = \underset{\mu}{\arg \min} \Big( \mathcal{J}(\mu) + \lambda \cdot \text{KL}(\mu \,||\, \mu^k) \Big).
\end{equation}
The analytical solution to this problem yields the closed-form Boltzmann update:
\begin{equation}
\label{eqn:boltzmann}
    \mu^*(x) \propto \mu^k(x) \cdot \exp\left( -\tfrac{1}{\lambda} f(x)\right).
\end{equation}
While $\mu^*$ is the optimal update, it is in general complex and difficult to sample from directly. 

% now we shift from general sequential to parametrized distributions
To achieve tractability, a common strategy is to constrain $\mu$ to a parametric family of distributions, typically a Gaussian $\,p_\theta = \mathcal{N}(\bar{x}, \Sigma)\,$, where $\bar{x}$ denotes the mean and $\Sigma$ the covariance. 
The objective then becomes to identify the finite-dimensional parameters $\theta$ such that $p_\theta$ best approximates the optimal Boltzmann distribution $\mu^*$ in \eqref{eqn:boltzmann}.
This particular instantiation of sequential KL control with a Gaussian parametrization underlies sampling-based algorithms such as Model Predictive Path Integral (MPPI) control~\cite{Theodorou2012RelativeEA, Williams2017ModelPP}, which have gained significant traction in the robotics community in recent years~\cite{howell2022predictivesamplingrealtimebehaviour,Turrisi2024OnTB,crestaz26mppi,dialmpic25,Zhai2025PAMPPIPM,schramm2026referencefreesamplingbasedmodelpredictive}.
These approaches typically rely solely on zeroth-order information (function evaluations). While such derivative-free formulations naturally handle non-smooth dynamics, they do not exploit available gradient or curvature information. As a result, these sampling-based methods often exhibit high variance and low sample efficiency in complex cost landscapes, in contrast to classical optimization techniques~—~such as Newton-based methods~—~that can achieve second-order convergence rates near a local optimum.

In the following, we extend this framework by embedding geometric information into the MPPI formulation, enabling the sampling process to exploit local curvature and thereby achieving faster, more reliable convergence.

\section{Methodology}
\label{sec:method}

Sampling-based control methods like MPPI offer substantial flexibility but often suffer from high variance and low sample efficiency.
To address these limitations, we propose a hybrid MPPI approach that incorporates geometric information into the sampling process via a local approximation of the objective function.
The key insight is to decompose the objective into an approximate model term and a residual term.
The resulting model-based information biases the sampling distribution toward informative regions, reducing the variance of the underlying gradient estimators and thereby accelerating convergence relative to the classic MPPI updates.
The remainder of this section details the formulation, the construction of the approximation, and stability mechanisms.

\subsection{Model-guided distribution update}

Consider the problem of finding the optimal distribution $p^*(x)$ at iteration $k+1$, given a prior estimate $p_{\theta^k}(x)$ obtained at iteration $k$. 
Following~\ref{eqn:boltzmann}, the information-theoretic optimal update takes the Boltzmann form:
\begin{equation}
\label{eqn:boltzmann-param}
    p^*(x) \propto p_{\theta^k}(x) \exp\left( - f(x) / \lambda \right).
\end{equation}
At iteration $k$, we decompose the objective function $f$ into a model $m^k$ and a residual term $r^k$, such that $f(x) = m^k(x) + r^k(x)$.
Here, $m^k$ acts as a control variate intended to capture the dominant landscape geometry, while $r^k$ accounts for unmodeled discrepancies.
Substituting this decomposition into~\eqref{eqn:boltzmann-param} yields the following factorization:
\begin{equation}
\label{eqn:factorize-general}
    p^*(x) \propto \underbrace{\left[ p_{\theta^k}(x) \exp\left( - m^k(x) / \lambda \right) \right]}_{\text{model-guided prior } \tilde{p}_{\theta^k}(x)} \cdot \exp\left( - r^k(x) / \lambda \right).
\end{equation}
Equation~\eqref{eqn:factorize-general} reveals a fundamental separation of concerns. The term in brackets, called model-guided prior, combines the previous prior with the explicit model to form a new intermediate distribution, which we denote as the \emph{model-guided prior} $\tilde{p}_{\theta^k}(x)$. 
Rather than sampling from an uninformative prior and weighing by the full complex objective $f$ as done classically in MPPI-based approaches, we sample from the structurally informed $\tilde{p}_{\theta^k}$ and weight only by the residual $r^k$.

Importantly, this formulation is valid for \emph{any} choice of model. This generality offers a powerful degree of freedom: we can design $m^k$ such that the resulting model-guided prior $\tilde{p}_{\theta^k}$ admits a closed-form solution. By selecting a model structure compatible with the prior, we ensure that the new sampling distribution remains analytically tractable.

\subsection{Closed-form model-guided prior via quadratic model approximation}
\label{subsec:closed-form-guided-prior}
\begin{small}
\begin{algorithm}[t]
    \DontPrintSemicolon
    \caption{\textbf{MPPI with Quadratic Model\\ Approximation}}
    \label{alg:mppi_quadratic}
    
    \SetKwInOut{Input}{Input}
    \Input{Initial mean $\bar{x}^0$ and covariance $\Sigma^0$, samples $N$, temperature $\lambda$}
    \BlankLine
    
    \For{$k \gets 0, 1, 2, \ldots$}{
        \tcp{1. Construct quadratic model analytically or via RS using Eq. \eqref{eq:rs_grad},~\eqref{eq:rs_hess}}
        Obtain gradient $g^k$ and hessian $H^k$ 
        
        \BlankLine
        \tcp{2. Compute guided prior using Eq.~\eqref{eqn:guided_cov},~\eqref{eqn:guided_mean}}
        Compute guided covariance $\tilde{\Sigma}^k$  and mean $\tilde{x}^k$\;
        
        \BlankLine
        \tcp{3. Sample from guided prior \& evaluate}
        \For{$i \gets 1$ \KwTo $N$}{
            Sample $x^{(i)} \sim \mathcal{N}(\tilde{x}^k, \tilde{\Sigma}^k)$\;
            Compute residual $r^k(x^{(i)})$\;
        }
        
        \BlankLine
        \tcp{4. Update using Eq.~\eqref{eqn:update_rule}}
        Compute weights $\tilde{w}^{(i)}$ and update mean $\bar{x}^{k+1}$ \;
    }
\end{algorithm}
\end{small}

To derive an efficient closed-form algorithm, we instantiate the general framework using Gaussian approximations.
We assume the current estimate is a Gaussian distribution $p_{\theta^k}(x) = \mathcal{N}(x \,|\, \bar{x}^k, \Sigma^k)$ parametrized by $\theta^k = \{ \bar{x}^k, \Sigma^k \}$.
We choose $m^k$ to be a quadratic expansion of the cost around the current mean $\bar{x}^k$:
{
\small
\begin{equation}
\label{eqn:quad-model}
    m^k(x) = f(\bar{x}^k) + (g^k)^\top (x-\bar{x}^k) + \frac{1}{2} (x-\bar{x}^k)^\top H^k (x-\bar{x}^k),
\end{equation}
}
where $g^k$ and $H^k$ denote the gradient and Hessian parameters at the expansion point, see Sec.~\ref{subsec:constructing-model} for details.
Since the product of a Gaussian PDF and the exponential of a quadratic function is itself a Gaussian, the model-guided prior $\tilde{p}_{\theta^k}$ derived in \eqref{eqn:factorize-general} remains in the Gaussian family. We can therefore solve for its parameters $\tilde{\theta}^k = \{ \tilde{x}^k, \tilde{\Sigma}^k \}$ analytically.
The guided prior $\tilde{p}_{\theta^k}(x) = \mathcal{N}(x \,|\, \tilde{x}^k, \tilde{\Sigma}^k)$ is characterized by the updated covariance:
\begin{equation}
\label{eqn:guided_cov}
    \tilde{\Sigma}^k = \left((\Sigma^k)^{-1} + \tfrac{1}{\lambda} H^k  \right)^{-1}.
\end{equation}
To guarantee a valid positive-definite covariance matrix $\tilde{\Sigma}^k$, appropriate regularization or convexification is applied to $H^k$ whenever it is indefinite.
Consequently, the mean is given by:
\begin{equation} 
\label{eqn:guided_mean}
    \tilde{x}^k = \tilde{\Sigma}^k \left( (\Sigma^k)^{-1} \bar{x}^k + \tfrac{1}{\lambda} \left( H^k \bar{x}^k - g^k \right) \right).
\end{equation}
The final optimal distribution is then obtained by reweighting this model-guided Gaussian $\tilde{p}_{\theta^k}$ with the residual:
\begin{equation}
\label{eqn:optimal-p-guided}
    p^*(x) \;\propto\; \underbrace{\mathcal{N}(x \,|\, \tilde{x}^k, \tilde{\Sigma}^k)}_{\tilde{p}_{\theta^k}(x)} \cdot \exp\!\left(- r^k(x) / \lambda \right).
\end{equation}
Interestingly, this reformulation naturally lends itself to applying MPPI on the residual function $r^k$ while sampling trajectories $\{x^{(i)}\}_{i=1}^N$ from the model-guided prior $\tilde{p}_{\theta^k}$. 
The update is now given by:
\begin{small}
\begin{equation}
\label{eqn:update_rule}
    \bar{x}^{k+1} \approx \sum_{i=1}^N \tilde{w}^{(i)} \, x^{(i)}, 
    \; \text{with } \tilde{w}^{(i)} = \frac{\exp\left(- r^k(x^{(i)}) / \lambda \right)}{\sum_{j=1}^N \exp\left(- r^k(x^{(j)}) / \lambda \right)}.
\end{equation}
\end{small}

\noindent Compared to vanilla MPPI, which samples directly from $p_\theta^k$, this model-guided approach fundamentally alters the update mechanism. 
First, the analytical shift to $\tilde{p}_{\theta^k}$ integrates geometry information, blending the prior mean $\bar{x}^k$ with a Newton-like step derived from the model $m^k$. 
Second, the sampling process is reserved for exploring the residual error $r^k$ around this new center, rather than exploring the full cost $f$. This procedure is summarized in Algorithm~\ref{alg:mppi_quadratic}.

\subsection{Quadratic model approximations}
\label{subsec:constructing-model}
\begin{figure}[t]
  \centering
  \includegraphics[width=\linewidth]{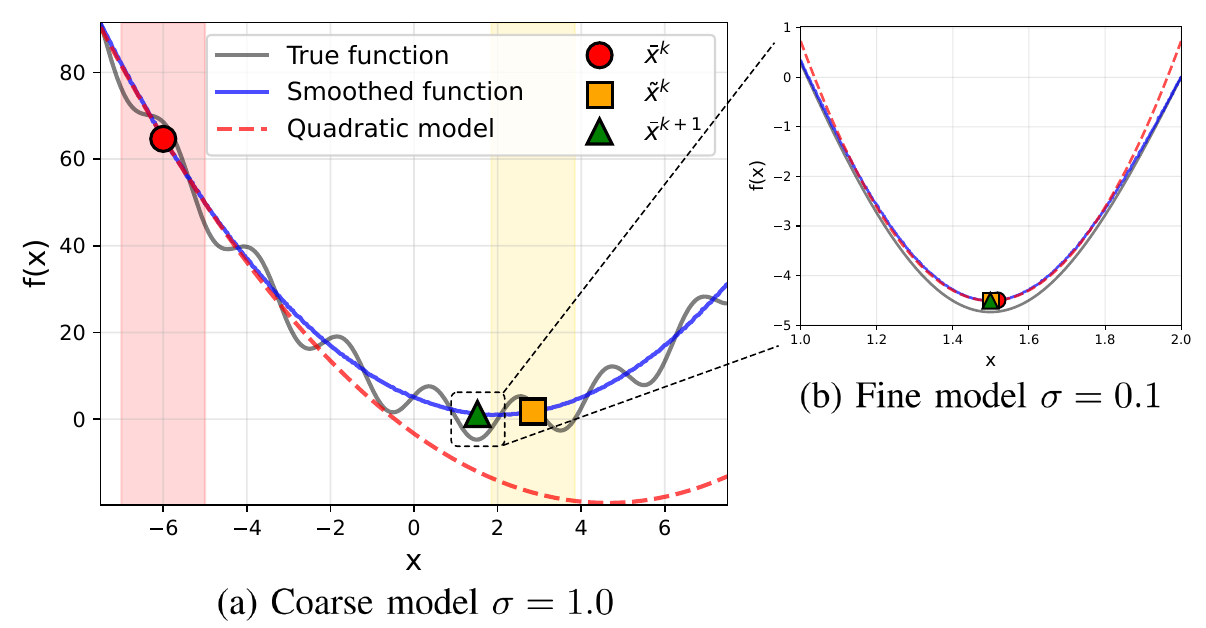}
  \caption{\textbf{Comparison coarse vs fine model.}
  (a) A large smoothing kernel $\sigma = 1.0$ suppresses sinusoidal disturbances (grey), yielding a coarse approximation (blue) capturing global geometry. The resulting quadratic model (red) generates a proposal $\tilde{x}^1$ moving toward the global minimum despite local nonconvexities. (b)~Near the optimum, reducing the scale to $\sigma = 0.1$ captures the local curvature for final convergence.}
    \label{fig:quadratic-sin}
    \vspace{-1.5em}
\end{figure}
To implement the update rules above, we require $g^k$ and $H^k$ to construct the model~\eqref{eqn:quad-model}. 
A strength of the proposed framework is that it decouples the sampling mechanism from the source of geometric information, allowing the practitioner to select the most appropriate approximation strategy based on the available computational budget and problem structure.\\

\noindent\textbf{Analytical second-order approximations.}
If $f$ is differentiable and smooth, $g^k$ and $H^k$ can be exact analytical derivatives. In this exact \emph{local model}, the resulting guided prior $\tilde{p}_{\theta^k}$ acts as a precise local Newton step. This maximizes convergence speed near the optimum but risks guiding the sampling distribution into shallow local minima if the current estimate is far from the global solution.\\

\noindent\textbf{Gauss-Newton approximations.}
Beyond exact derivatives, our framework accommodates strategies that exploit specific problem structures. For instance, many control objectives are formulated as non-linear least-squares problems: $f(x) = \frac{1}{2} \|R(x)\|^2$, where residual $R(x)$ stacks error terms such as state deviations and control penalties. In such cases, the Hessian can be approximated via the Gauss-Newton approximation as $H^k \approx J(x)^\top J(x)$, where $J(x)$ is the Jacobian of $R(x)$. This avoids computing the full Hessian while ensuring a positive-semidefinite approximation that seamlessly integrates into our quadratic model update.
Recent work~\cite{homburger2026gaussnewtonacceleratedmppicontrol} can be viewed as an instantiation of this principle, in which curvature information is reconstructed from black-box residuals via randomized smoothing.
\\

\noindent\textbf{Quasi-Newton approximations.}
Similarly, when second-order information is prohibitively expensive or unavailable, one can employ Quasi-Newton methods such as BFGS or L-BFGS~\cite{Nocedal2018NumericalO}. These techniques iteratively build a curvature estimate $H^k$ from only a history of gradient updates, enabling our model-guided framework to be applied in settings where only first-order derivatives are available.\\

\noindent\textbf{Randomized smoothing approximations.}
In the most general case where $f$ is non-smooth or black-box, we employ Randomized Smoothing (RS)~\cite{berthet20rs,duchi12rs,Abernethy16rs} to estimate the model parameters. This allows us to recover zeroth-order estimators for the gradient and Hessian using only function evaluations. While RS is a general framework compatible with various smoothing distributions, we use a Gaussian kernel in this work. This defines the smoothed surrogate as the Gaussian convolution $f_\sigma(x) = \mathbb{E}_{z \sim \mathcal{N}(0, \sigma^2 I)} [f(x + z)]$. 
By exploiting Stein's identity and utilizing a centered control variate to reduce variance, we obtain the following Monte Carlo estimators for the gradient $g^k$ and Hessian $H^k$ at the current mean $\bar{x}^k$:
\begin{align}
    \label{eq:rs_grad}
    g^k &\approx \frac{1}{M \sigma^2} \sum_{j=1}^M \left( f(\bar{x}^k + z_j) - f(\bar{x}^k) \right) z_j, \\
    \label{eq:rs_hess}
    H^k &\approx \frac{1}{M \sigma^4} \sum_{j=1}^M \left( f(\bar{x}^k + z_j) - f(\bar{x}^k) \right) (z_j z_j^\top - \sigma^2 I),
\end{align}
where $\{z_j\}_{j=1}^M$ are samples drawn from $\mathcal{N}(0, \sigma^2 I)$. 

With a small noise scale ($\sigma \to 0$), the smoothed surrogate $f_\sigma$ closely approximates the original objective $f$. The estimators~\eqref{eq:rs_grad},~\eqref{eq:rs_hess} capture fine-scale curvature information, behaving similarly to analytical derivatives. We refer to this as the \emph{fine model} (Fig.~\ref{fig:quadratic-sin}b). 
As $\sigma$ increases, RS aggregates objective information over larger neighborhoods around the current iterate. The gradient and Hessian are based on global structure rather than local details, allowing the algorithm to step over small obstacles that might trap a local optimizer. We refer to this as the \emph{coarse model} (Fig.~\ref{fig:quadratic-sin}a). The induced quadratic approximation biases the guided prior toward regions that appear promising under wider exploratory variations, encouraging exploration while filtering out fine local structure.

\subsection{Incremental update formulation}

Although formulated above in terms of full iterations, the update mechanism can equivalently be expressed in terms of incremental steps.
Consider a perturbation $\epsilon$ such that \mbox{$x = \bar{x}^k + \epsilon$}. A second-order expansion of the objective yields
{\small\begin{equation}
    f(\bar{x}^k + \epsilon) = f(\bar{x}^k) + (g^k)^\top \epsilon + \frac{1}{2}\epsilon^\top H^k \epsilon + r^k(\bar{x}^k + \epsilon),
\end{equation}}%
where the quadratic terms define the local model $m^k$ and $r^k$ accounts for unmodeled effects.
As in Sec.~\ref{subsec:closed-form-guided-prior}, combining the quadratic model with the $k$-th Gaussian exploration prior $p(\epsilon)=\mathcal{N}(0,\Sigma_{0k})$ induces a guided Gaussian distribution
\begin{equation}
    \tilde{p}_{\theta^k}(\epsilon) = \mathcal{N}(\epsilon \,|\, \tilde{\delta}^k, \tilde{\Sigma}^k),
\end{equation}
with parameters given by
\begin{align}
\label{eq:guided_step_cov}
    \tilde{\Sigma}^k &= \left(\Sigma_{0k}^{-1} + \tfrac{1}{\lambda}H^k\right)^{-1}, \\
\label{eq:guided_step_mean}
    \tilde{\delta}^k &= -(\lambda\Sigma_{0k}^{-1}+H^k)^{-1} g^k,
\end{align}
where $\tilde{\delta}^k$ represents the model-guided incremental update step mean.
As before, we regularize $H^k$ if necessary to ensure $\tilde{\Sigma}^k$ is positive-definite. 
The optimal update step distribution is then obtained by reweighting this guided Gaussian by the residual,
\begin{equation}
    p^*(\epsilon) \propto \tilde{p}_{\theta^k}(\epsilon) \exp\!\left(-r^k(\bar{x}^k+\epsilon) / \lambda \right).
\end{equation}
This is the direct update-step analogue of the iterate-level distribution in~\eqref{eqn:optimal-p-guided}. Yet, this formulation choice exposes the prior covariance $\Sigma_{0k}$ as a tunable parameter in the algorithm design, a feature that will be exploited in the next section~\ref{subsec:variance_control} to adequately control the variance.

A natural strategy is to perform the parameter update along the mean of this optimal distribution:
\begin{equation}
    \Delta\bar{x}^k = \mathbb{E}_{p^*}[\epsilon].
\end{equation}
Since $p^*$ is available only up to a normalizing constant, we estimate this expectation via self-normalized importance sampling using $\tilde{p}_{\theta^k}$ as the proposal. We first express the expectation as a ratio:
\begin{equation}
    \Delta\bar{x}^k = \frac{ \mathbb{E}_{\tilde{p}_{\theta^k}}\!\left[ \epsilon\,\exp\!\left(-r^k(\bar{x}^k+\epsilon) / \lambda \right) \right] }{ \mathbb{E}_{\tilde{p}_{\theta^k}}\!\left[ \exp\!\left(- r^k(\bar{x}^k+\epsilon) / \lambda \right) \right] }.
\end{equation}
Drawing samples $\{\epsilon_i\}_{i=1}^N \sim \mathcal{N}(\tilde{\delta}^k, \tilde{\Sigma}^k)$ from the guided prior, we obtain the Monte Carlo approximation:
{\small
\begin{equation}
    \Delta\bar{x}^k \approx \sum_{i=1}^N w_i\,\epsilon_i, \; \text{with} \; w_i = \frac{ \exp\!\left(-r^k(\bar{x}^k+\epsilon_i) / \lambda \right) }{ \sum_{j=1}^N \exp\!\left(-r^k(\bar{x}^k+\epsilon_j) / \lambda \right) }.
\end{equation}}%
\noindent Finally, the mean parameter is updated as:
\begin{equation}
    \bar{x}^{k+1} \leftarrow \bar{x}^k + \Delta\bar{x}^k.
\end{equation}

\subsection{Stability and variance control of the model-guided prior}
\label{subsec:variance_control}

While the update rules in \eqref{eqn:guided_cov} and \eqref{eq:guided_step_cov} directly incorporate curvature, they can lead to variance collapse since they can only decrease variance. In regions where the Hessian $H^k$ has large positive eigenvalues, the resulting guided covariance $\tilde{\Sigma}^k$ may shrink excessively, hindering the exploration required to resolve the residual $r^k$. To mitigate this, we use regularization strategies that ensure stability and maintain a minimum exploration width.\\

\noindent\textbf{Polyak/Juditsky averaging.} 
To smooth the evolution of the sampling distribution and prevent abrupt jumps caused by noisy gradient or Hessian estimates, we apply exponential moving averages to the model-guided parameters. Instead of using the raw guided mean $\tilde{\delta}^k$ and covariance $\tilde{\Sigma}^k$ directly for the current step, we update the sampling parameters via an exponential moving average with smoothing factor $\alpha_\delta, \alpha_\Sigma \in [0, 1]$:
\begin{align}
\bar{\delta}^k &= \alpha_\delta \tilde{\delta}^k + (1-\alpha_\delta) \bar{\delta}^{k-1}, \\ \bar{\Sigma}^k &= \alpha_\Sigma \tilde{\Sigma}^k + (1-\alpha_\Sigma) \bar{\Sigma}^{k-1}.
\end{align}

\noindent\textbf{Variance control.} 
While the above averaging acts as a temporal filter, it does not guarantee a lower bound on the variance. To strictly prevent collapse, we use a scaling strategy that regulates the distribution's magnitude while preserving its geometry. By applying a scalar gain, we maintain the anisotropic structure (eigenvectors and relative aspect ratios) derived from the Hessian, ensuring the sampling ellipsoid remains aligned with the landscape curvature even when inflated to a safe minimum volume.

We derive the required input noise $\sigma_{0k}^2$ by analyzing the covariance update in the principal directions of the local curvature. Starting from the update rule in Eq.~\eqref{eq:guided_step_cov} and assuming an isotropic prior covariance $\Sigma_{0k} = \sigma_{0k}^2 I$, the inverse of the updated covariance becomes:
\begin{equation}
(\tilde{\Sigma}^k)^{-1} = \frac{1}{\sigma_{0k}^2} I + \frac{1}{\lambda} H^k.
\end{equation}
Since $H^k$ is symmetric, we can analyze the update along its principal axes. Let $\kappa_i$ denote the eigenvalues of $H^k$. The updated covariance $\tilde{\Sigma}^k$ shares same eigenbasis, with its eigenvalues $\tilde{\sigma}_i^2$ given by the reciprocal sum:
\begin{equation}
    \frac{1}{\tilde{\sigma}_{i}^2} = \frac{1}{\sigma_{0k}^2} + \frac{\kappa_i}{\lambda} \quad \Rightarrow \quad \tilde{\sigma}_{i}^2 = \frac{\lambda \, \sigma_{0k}^2}{\lambda + \sigma_{0k}^2 \, \kappa_i}.
\end{equation}
We see that the variance is most compressed in the direction of maximum positive curvature. 
To prevent excessive narrowing, we enforce that the variance in this "worst-case" direction, associated with the maximum eigenvalue $\kappa_{\max} (H^k)$, must be at least a target threshold $\sigma_{\text{target}}^2$:
\begin{equation}
\frac{\lambda \, \sigma_{0k}^2}{\lambda + \sigma_{0k}^2 \, \kappa_{\max}} \geq \sigma_{\text{target}}^2,
\end{equation}
yielding the lower bound:
\begin{equation}
    \sigma_{0k}^2 \geq \frac{\lambda \, \sigma_{\text{target}}^2}{\lambda - \sigma_{\text{target}}^2 \, \kappa_{\text{max}}}.
\end{equation}
This ensures that the sampling distribution maintains a desired width in the most constrained direction, thereby guaranteeing that variance in all other directions exceeds the threshold without distorting the curvature information.

\section{Experiments}
\label{sec:experiments}

We investigate how incorporating a quadratic model into MPPI affects convergence speed, solution quality, and robustness in low-sample regimes. In particular, we study how the \emph{locality} and \emph{fidelity} of the quadratic model approximation $m^k$ influence performance.
To isolate specific algorithmic properties, we consider problems with increasing difficulty. We begin with static optimization benchmarks to validate the core optimization mechanism. Then, we examine continuous control in the context of cart-pole control, a smooth but underactuated, nonlinear dynamical system, exploiting analytical derivatives and assessing the performance of Gauss-Newton and Quasi-Newton approximations. Finally, we aim to evaluate robustness to non-smooth dynamics (Single-Finger Sphere Manipulation) in contact-rich scenarios using randomized smoothing. 

As baselines, we consider 
(i) vanilla MPPI, which samples from a fixed Gaussian prior and evaluates trajectories using the objective $f$, without exploiting any explicit local structure, and 
(ii) CMA-ES, implemented using the \texttt{pycma} Python library~\cite{hansen2019pycma}, a state-of-the-art derivative-free optimizer that adapts a full covariance matrix. 
We compare this against several model-guided MPPI variants that adopt the incremental update formulation and differ only in how the quadratic approximation in~\eqref{eqn:quad-model} is constructed.

Our experiments highlight three different key effects of model guidance: 
(i)~accelerated and more accurate convergence when reliable analytical information is available, 
(ii)~increased robustness in low-sample regimes, where purely sampling-based methods suffer from weight degeneracy, and
(iii)~the complementary roles of coarse and fine quadratic models in capturing global structure versus local precision. 

\subsection{Illustration of quadratic model guidance}

We first show the mechanism on a 2D narrow-valley objective illustrated in Fig.~\ref{fig:quadratic-narrow-valley}. 
This visualizes how second-order information reshapes the sampling distribution to align with the cost geometry. By effectively performing a Newton-like update for the distribution mean, Model-Guided MPPI concentrates samples in high-probability regions. 

To evaluate sample quality, we use the effective sample size (ESS)~\cite{martine2016ess}, a standard diagnostic for importance sampling that quantifies the weight degeneracy induced by the sampling distribution.
Given normalized importance weights $w_i$, the ESS is defined as:
\begin{equation}
    w_i = \frac{\exp(-f_i / \lambda)}{\sum_j \exp(-f_j / \lambda)}, \quad \mathrm{ESS} = \frac{1}{\sum_i w_i^2}.
    \label{eqn:ESS}
\end{equation}
The ESS measures the number of effectively contributing samples; larger values indicate lower variance in the importance weights. In the context of MPPI, a higher ESS corresponds to more informative samples.
Using the same parameters ($N=100$, $\lambda=0.1$, $\sigma=0.2$) in Fig.~\ref{fig:quadratic-narrow-valley}, vanilla MPPI collapses to an ESS of $1.1$, indicating severe weight degeneracy. In contrast, our method maintains an ESS of $23.4$, demonstrating efficient coverage of the relevant state space.

\begin{table}[t]
    \centering
    \setlength{\tabcolsep}{2.5pt}
    \definecolor{lightgraytext}{gray}{0.6}
    \rowcolors{1}{white}{pastelblue}
    \footnotesize
    \begin{tabular}{l|rrr} 
        \toprule
        Function & \multicolumn{1}{c}{Model-Guided MPPI (ours)} & \multicolumn{1}{c}{Vanilla MPPI~~} & \multicolumn{1}{c}{CMA-ES} \\
        \midrule
        Rosenbrock      & $6.9 \pm 2.5$ \textcolor{lightgraytext}{\textemdash}~~~~~~~\;   & $21.4 \pm 7.3$ \textcolor{lightgraytext}{\textemdash}~\,  & $12.0 \pm 1.3$ \textcolor{lightgraytext}{\textemdash} \\
        Styblinski–Tang & $2.1 \pm 0.4$ \textcolor{lightgraytext}{\textemdash}~~~~~~~\;     & $9.2 \pm 0.8$ {\tiny $(14)$}& $7.8 \pm 2.9$ \textcolor{lightgraytext}{\textemdash} \\
        Rastrigin & $3.0 \pm 1.1$ \textcolor{lightgraytext}{\textemdash}~~~~~~~\;     & $6.9 \pm 6.4$ \textcolor{lightgraytext}{\textemdash}~\,   & $5.5 \pm 4.6$ \textcolor{lightgraytext}{\textemdash} \\
        Ackley & $3.3 \pm 1.6$ {\scriptsize $(1)$}~~~~~~~   & $3.8 \pm 2.1$ \textcolor{lightgraytext}{\textemdash}~\,   & $3.5 \pm 1.4$ \textcolor{lightgraytext}{\textemdash} \\
        \bottomrule
    \end{tabular}
    \caption{Mean iterations $\pm$ std (failures) over 100 seeds.}
    \label{tab:results-benchmark}
    \vspace{-2em}
\end{table}

\subsection{Performance with analytical gradients}
We first validate the method in a setting where exact gradient and Hessian information are available. 
Here, the quadratic model $m^k$ accurately captures local curvature for smooth functions, allowing us to isolate and analyze the benefits of the update rule from errors in gradient estimation. \\

\noindent\textbf{Static benchmarks.} Tab.~\ref{tab:results-benchmark} reports the number of iterations required by each method to reach the global optimum within a specified infinity-norm distance using $100$ samples. Initializations were set to $\bar{x}^0 = (0,0)$ (Rosenbrock, Styblinski–Tang), $(2.0, 2.0)$ (Ackley), and $(1.9, 1.7)$~(Rastrigin). 
The results confirm the efficacy of the model guidance: our framework consistently converges faster than the baselines and, crucially, exhibits lower variance across the 100 random seeds. For instance, on the Rastrigin benchmark, we achieve a tight convergence distribution ($3.0 \pm 1.1$ iterations), whereas Vanilla MPPI ($6.9 \pm 6.4$) and CMA-ES ($5.5 \pm 4.6$) display higher volatility, reflecting their sensitivity to the stochasticity of the sampling process. \\

\noindent\textbf{Continuous control.} 
To assess our method on a continuous control problem, we consider a nonlinear, underactuated cart-pole swing-up task with a horizon length of $2.5\,\mathrm{s}$ and a time step of  $50\,\mathrm{ms}$. The objective is formulated as a finite-horizon trajectory optimization problem over a sequence of continuous controls.
\begin{figure}[t]
    \centering
    \begin{subfigure}[b]{1.0\linewidth}
        \centering
        \includegraphics[width=\linewidth]{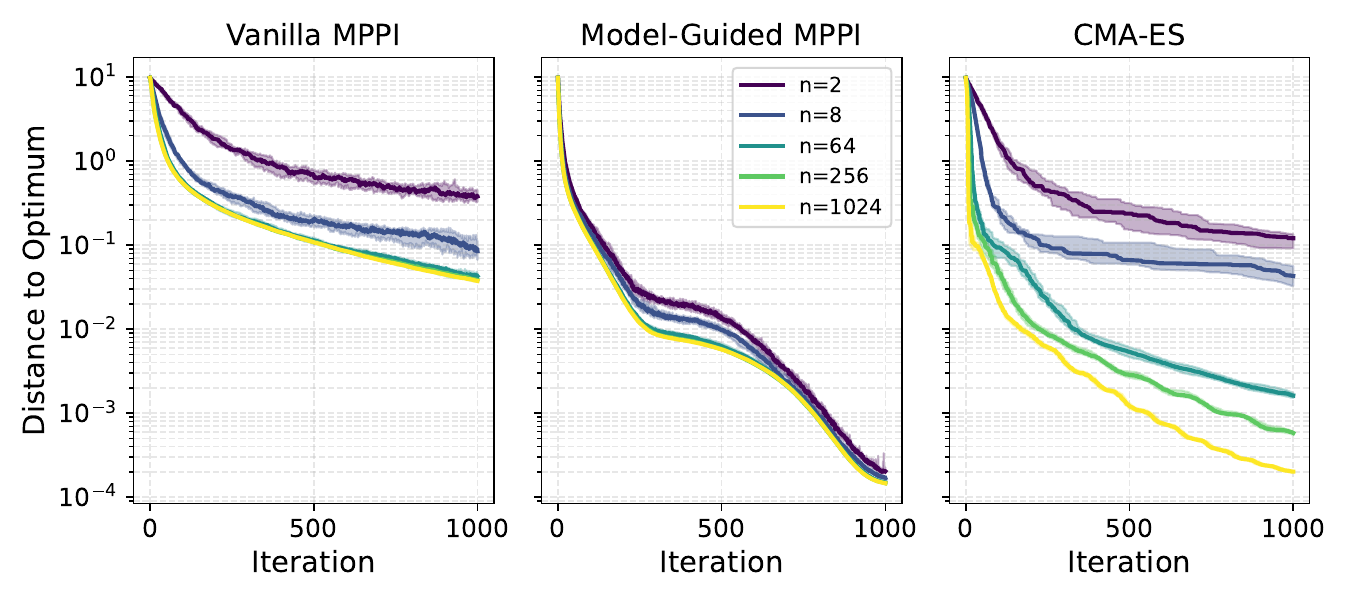}
        \caption{\textbf{Convergence speed.} Median distance to the Newton-optimal solution over iterations. Shaded areas indicate the interquartile range (IQR) over 20 seeds.}
        \label{fig:cartpole-convergence}
    \end{subfigure}    
    \vspace{1em}
    \begin{subfigure}[b]{1.0\linewidth}
        \centering
        \includegraphics[width=\linewidth]{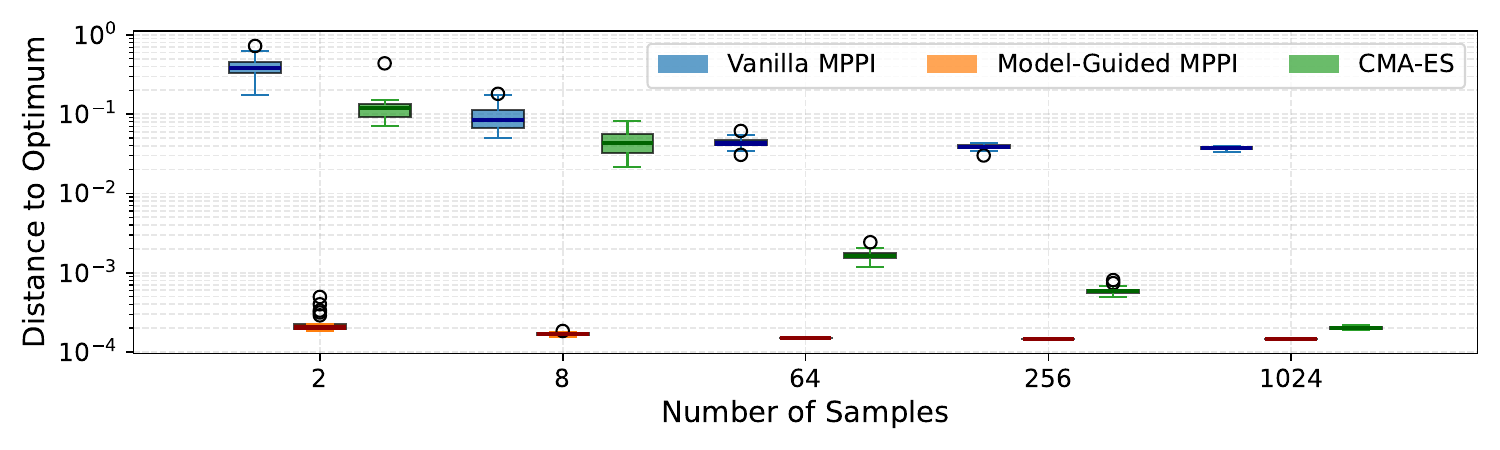}
        \caption{\textbf{Variance analysis in low-sample regime.} Box-plot of the final distance to the optimum after $1000$ iterations. The log-scale highlights the variance and error increase for baselines when $N < 64$.}
        \label{fig:cartpole-variance}
    \vspace{-1em}
    \end{subfigure}
    \caption{\textbf{Cart-pole swing-up results}. Comparison of convergence behavior and variance across sample budgets \mbox{$N \in \{2, \dots, 1024\}$} and $20$ random seeds.} 
    \label{fig:cartpole-combined}
    \vspace{-1em}
\end{figure}
We leverage open-source tools to implement the cart-pole dynamics and cost function using the Pinocchio library~\cite{carpentier2019pinocchio, pinocchioweb}, and CasADi~\cite{Andersson2019}. This provides access to exact analytical expressions for the trajectory cost, its gradient, and Hessian with respect to the control sequence through automatic differentiation. As a reference for the true optimum over the given horizon and discretization, we compute a locally optimal control sequence using a Newton solver with a line search.

We evaluate convergence behavior across a wide range of sampling budgets $N \in \{2, \dots, 1024\}$ in Fig.~\ref{fig:cartpole-combined}. In the high-sample regime ($N=1024$), both CMA-ES and Model-Guided MPPI converge rapidly to the optimum ($<10^{-4}$ error), whereas Vanilla MPPI requires significantly more iterations.
However, the baselines are sensitive as the number of samples is reduced. Vanilla MPPI exhibits increasingly slow convergence and high variance, while CMA-ES becomes unstable below $N=64$. In contrast, our method demonstrates sample invariance by leveraging an analytical quadratic model to effectively decouple performance from the sampling budget. As shown in Fig.~\ref{fig:cartpole-convergence}, the convergence trajectories from $N=2$ to $1024$ collapse onto a single curve. Fig.~\ref{fig:cartpole-variance} further quantifies this robustness: while the baselines suffer from large variance in the low-sample regime, Model-Guided MPPI maintains negligible variance and consistent high-precision convergence even with as few as $N=2$ samples. This confirms that when reliable second-order information is available, the quadratic model successfully guides optimization, reducing reliance on massive parallel rollouts. \\

\noindent\textbf{Hessian approximations.} While the previous results leverage exact analytical Hessians, we now investigate the impact of approximation quality. Following the strategies in Sec.~\ref {subsec:constructing-model}, we compare the exact local model against Gauss-Newton (GN), iterative BFGS approximations, and an Adam-based approximation (interpreting the second moment as a diagonal Hessian), alongside a vanilla MPPI baseline on the same cart-pole task using $8$ samples.
\begin{figure}
    \centering
    \includegraphics[width=1\linewidth]{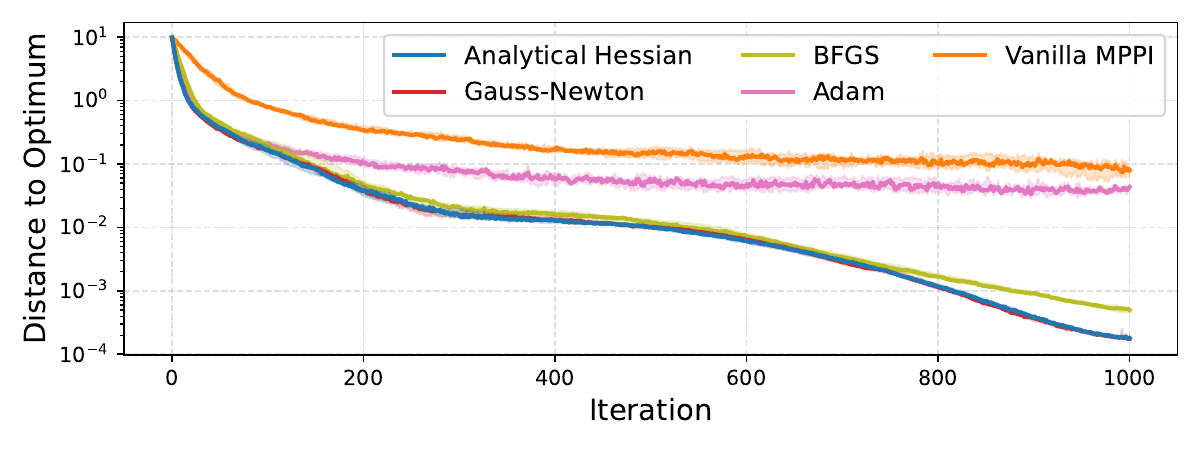}
    \caption{\textbf{Convergence comparison of Hessian approximations on the cart-pole swing-up task.} We compare dense curvature models (Analytical, Gauss-Newton, BFGS), against a diagonal approximation (Adam) and the isotropic baseline (Vanilla MPPI). The median distance to the reference optimum is plotted across 10 random seeds, with shaded areas indicating the interquartile range (IQR).}
    \label{fig:hessian-comparison}
    \vspace{-1em}
\end{figure}
As shown in Fig.~\ref{fig:hessian-comparison}, the fidelity of the curvature information is critical. While the Adam-based model-guidance improves upon the baseline, its diagonal approximation proves insufficient to achieve high precision.
In contrast, variants that capture dense curvature information, such as exact, GN, or BFGS, achieve high-precision convergence.
Notably, the GN approximation exploits the least-squares structure of the cost and closely matches the analytical Hessian. 
The BFGS variant, despite relying solely on iterative gradient updates, maintains a similar trajectory for most of the optimization, with only a minor performance gap emerging as it approaches the higher-precision regime ($< 10^{-3}$).

\subsection{Coarse and fine quadratic model approximations}

We next evaluate how the locality of the quadratic model, controlled by the smoothing scale, affects performance on nonconvex objectives. To build intuition, we consider a one-dimensional illustrative example in Fig.~\ref{fig:quadratic-sin}, in which the convex objective is corrupted by sinusoidal noise. Here, the trade-off becomes visible: a coarse model (large $\sigma$) filters out high-frequency perturbations to recover the dominant global geometry. This prevents the optimizer from being trapped in local minima and guides exploration toward the correct basin of attraction. Once the iterate is in the vicinity of the optimum, a fine model (small $\sigma$) is used to resolve accurate local curvature for high-precision convergence.

This intuition extends to higher-dimensional challenges, such as the Rastrigin function with initial guess $(1.9, 1.7)$. By starting with a large noise kernel ($\sigma = 1.5$), Model-Guided MPPI discovers the global basin of attraction around $(0,0)$. Transitioning to a finer kernel ($\sigma=0.1$), then employs a Newton-like update step to achieve high-precision optimization in just 3 iterations with $100$ samples.

\subsection{Randomized smoothing on non-smooth dynamics}
\begin{figure}[t]
    \centering
    \includegraphics[width=\linewidth]{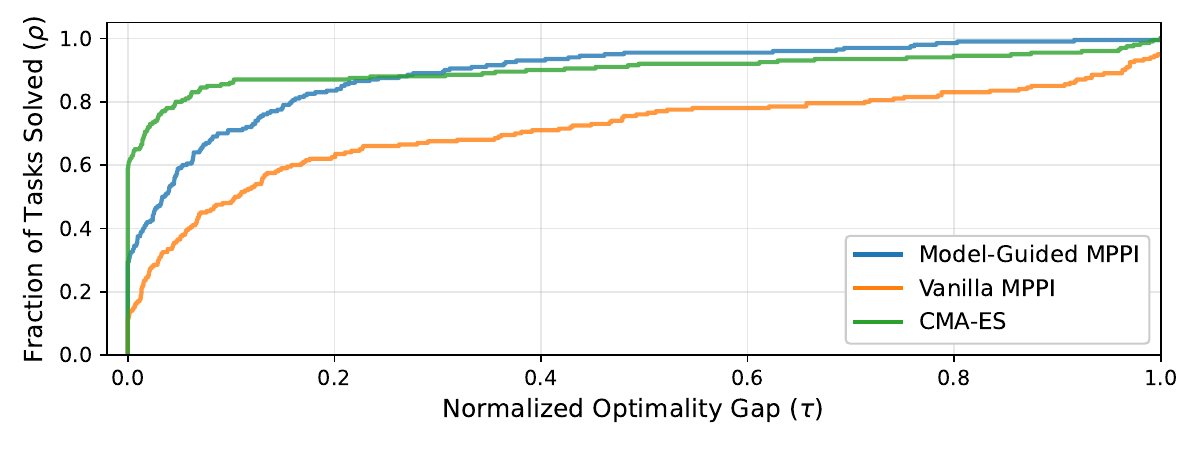}
    \caption{\textbf{Performance profile on Single-Finger Sphere Manipulation.} Comparison across $200$ randomized tasks.}
    \label{fig:perf-profiles}
\end{figure}
\begin{figure}[t]
    \centering
    \includegraphics[width=0.9\linewidth]{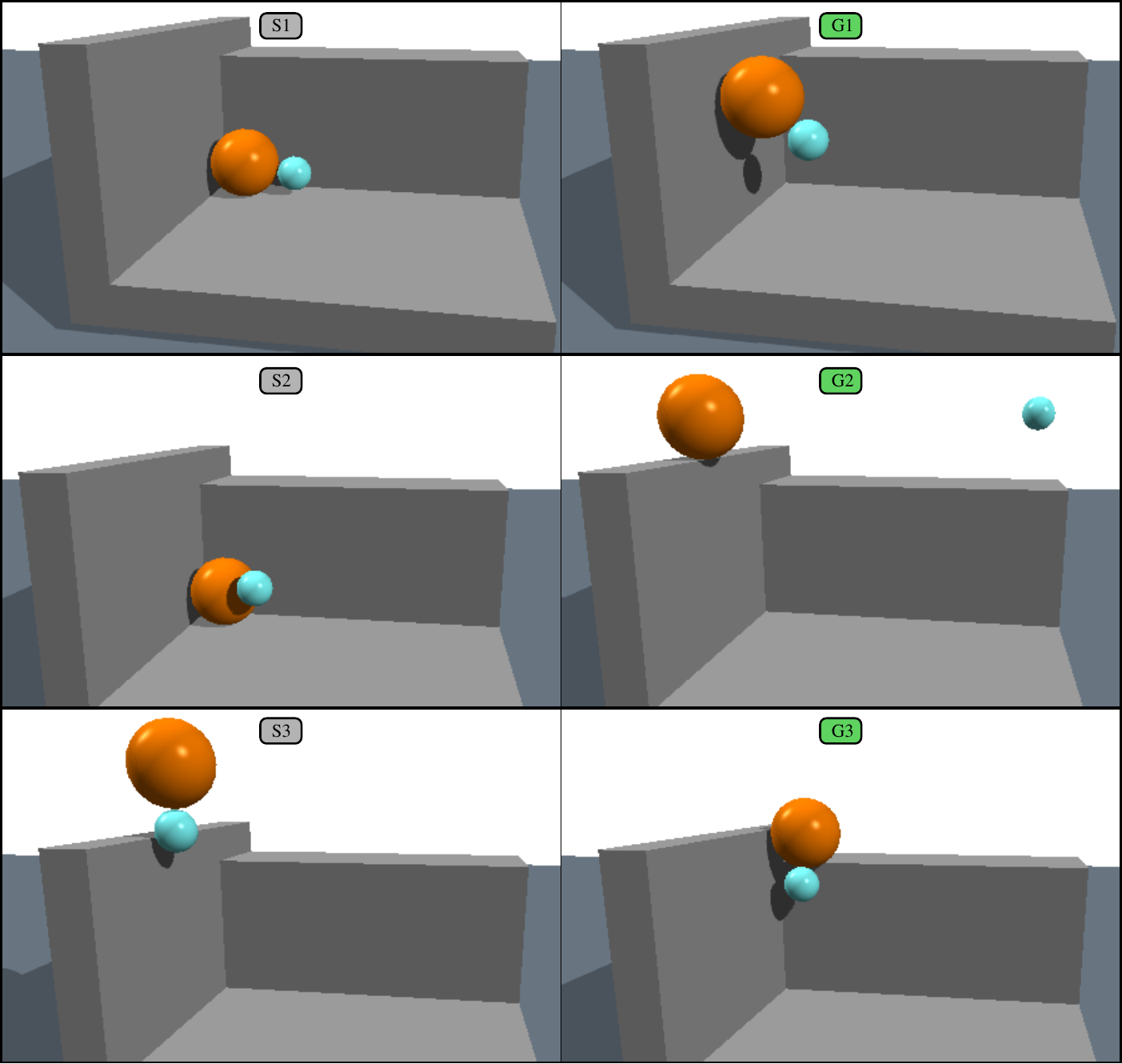}
    \caption{\textbf{Single-Finger Sphere Manipulation.} Three different starting states (left) and corresponding goal states (right) obtained via NLP sampling illustrate varying degrees of task difficulty. The task requires the actuated finger (blue) to navigate from its start to a target configuration, while simultaneously manipulating the passive object (orange) from its initial position to a specified goal pose.}
    \label{fig:spheres-benchmark}
    \vspace{-1em}
\end{figure}
To demonstrate the versatility of our framework, we apply it to a challenging contact-rich manipulation task. Contact dynamics introduce discontinuities that make standard analytical gradients uninformative. In this setting, we use randomized smoothing to extract quadratic model-guidance for a black-box physics simulator. \\

\noindent\textbf{Benchmark task.} We use a single-finger and sphere manipulation scenario as a minimal proxy for contact-rich manipulation. The system comprises two spheres in a static environment with walls, as shown in Fig.~\ref{fig:spheres-benchmark}. 
While geometrically simple, this task captures the core complexity of manipulation planning: non-smooth dynamics and underactuation. The actuated "finger" (blue sphere) must navigate to a target while manipulating a passive object (orange sphere), requiring the optimizer to manage frequent contact-making and breaking to transfer momentum. \\

\noindent\textbf{Experimental setup.} This task is implemented within the \texttt{robotic} library~\cite{Toussaint2024Robotic}, a framework for manipulation planning and constrained optimization which interfaces with the \texttt{MuJoCo} physics engine~\cite{todorov2012mujoco}. The simulation handles system dynamics, including friction and contact impulses at a timestep of $1\text{ms}$.
We parameterize the robot's motion as a spline trajectory defined by 4 control points over a fixed horizon of $1.0\text{s}$, where the optimization variables are the spline control points. \\ 

\noindent\textbf{Evaluation protocol.}
We generate $200$ diverse problem instances with challenging contact configurations via NLP sampling~\cite{Toussaint2024NLPSC} and compare methods using performance profiles of the normalized optimality gap after a fixed budget of $30$ iterations. We enforce a fixed planning budget of $64$ trajectory samples per iteration for all methods.
For our model-guided approach, we use randomized smoothing with $128$ auxiliary samples to estimate the quadratic model approximation. We treat this estimation step as a surrogate for an analytical gradient oracle. It represents the cost of obtaining model information in a non-differentiable setting, while the planning update itself respects the same sample constraint as the baselines. \\

\noindent\textbf{Metric and results.}
Since the global optimum is unknown, we define the best solution $f^*$ as the minimum cost found by any method for a given task. The normalized optimality gap is then defined as $\tau = (f_\text{final} - f^*) / (f_\text{init} - f^*)$. A value of $\tau=0$ indicates that the method matched the best-performing method, while $\tau=1$ implies no improvement.

The performance profile, depicted in Fig.~\ref{fig:perf-profiles}, reports the fraction of tasks where a method achieves a solution within a threshold $\tau$ of the best performance.
Model-Guided MPPI consistently achieves low normalized optimality gaps across a large fraction of problem instances, indicating reliable convergence, whereas Vanilla MPPI exhibits higher variability. CMA-ES demonstrates competitive performance in most instances but shows slightly reduced consistency across the full task distribution. 
While this benchmark serves as a first step, the results confirm that the proposed formulation can handle the hybrid dynamics inherent to rich-contact manipulation scenarios. This suggests a promising path toward applying variance-reduced MPPI to full-scale dexterous manipulation tasks in future work.

\subsection{Computational analysis}
\label{subsec:computational-analysis}
Finally, we examine how model construction and residual sampling contribute to the per-iteration cost across our three setups. Vanilla MPPI incurs the same sampling cost but evaluates the full objective $f$ rather than the residual $r^k$, so the overhead of our method is entirely concentrated in model construction, whose magnitude depends on the chosen approximation strategy.

\begin{table}[t]
\centering
% \captionsetup{font=footnotesize}
\caption{Per-iteration timings $\pm$ std over 10 seeds using $N$ samples and percentages denote share of total time. Static benchmark (SB) uses analytical derivatives, cart-pole (CP) uses AD-based exact derivatives and non-smooth single-finger (SF) manipulation uses randomized smoothing.}
\label{tab:timing}
\scriptsize
\setlength{\aboverulesep}{0pt}
\setlength{\belowrulesep}{0pt}
\rowcolors{3}{pastelblue}{white}
\setlength{\tabcolsep}{3pt}
\begin{tabular}{l|c@{\hspace{1.1\tabcolsep}}r|c@{\hspace{1.1\tabcolsep}}r}
\toprule
Task\;-\;$(N)$ & Model Construction [ms] & & Sampling $r^k$ or $f$ [ms] & \\
\midrule
SB\;-\;$(100)$  & $0.2 \pm 0.0$& (12\%) & $1.4 \pm 0.2$ &(88\%) \\
\midrule
CP\;-\;$(2)$    & $3.7 \pm 0.5$ &(75\%) & $1.2 \pm 1.8$ &(25\%) \\
CP\;-\;$(8)$    & --- &(70\%) & $1.5 \pm 3.5$ &(30\%) \\
CP\;-\;$(64)$   & --- &(52\%) & $3.6 \pm 2.1$ &(48\%) \\
CP\;-\;$(256)$  & --- &(26\%) & $10.8 \pm 3.0~\,$ &(74\%) \\
CP\;-\;$(1024)$ & --- &(\phantom{0}9\%) & $39.5 \pm 4.5~\,$ &(91\%) \\
\midrule
SF\;-\;$(64)$   & $1412.1 \pm 156.6~$ &(68\%) & $706.4 \pm 78.1~\,$ &(32\%) \\
\bottomrule
\end{tabular}
\vspace{-0.7em}
\end{table}

Tab.~\ref{tab:timing} reports per-iteration timings, measured on an Intel Core i7-11850H. (i) Static benchmarks (SB): model construction is negligible ($12\%$ vs.\ $88\%$) since derivatives are available in closed form. The timings are averaged over the four functions. 
(ii) Cart-pole (CP): using AD-based exact derivatives, model construction cost is constant and independent of the sample budget $N$, while residual sampling grows linearly. The model share consequently decreases from $75\%$ at $N=2$ to $9\%$ at $N=1024$. We use AD here to validate the variance-reduction principle under an exact model, not for wall-clock efficiency. 
(iii) Single-finger (SF): model construction dominates ($68\%$ vs.\ $32\%$), since we use $128$ auxiliary samples (twice the planning budget) for randomized smoothing.

Overall, the method is most attractive when the quadratic model is already available or cheap relative to rollouts. When it must be estimated from scratch, the trade-off becomes a balance between the additional estimation cost and the gain in sample efficiency. The framework remains agnostic to this choice, supporting cheaper structured approximations such as Gauss--Newton or BFGS that closely track the exact model (Fig.~\ref{fig:hessian-comparison}) at a fraction of the cost.

\section{Discussion and Conclusion}
\label{sec:conclusion}

In this work, we introduced a variance-reduced, accelerated extension of MPPI based on a model–residual decomposition of the objective. By factoring the Boltzmann update into a model-guided prior and a residual correction, the method injects structural information into the sampling process while retaining the flexibility of sampling-based control.
When instantiated with a quadratic model, this decomposition yields a closed-form Gaussian guided prior whose mean and covariance encode first- and second-order information, resulting in a Newton-like update at the distribution level. A key strength is that the framework is agnostic to the source of curvature information, seamlessly accommodating analytical Hessians, structural approximations (e.g., Gauss- or Quasi-Newton), and gradient-free randomized smoothing.
Across benchmark optimization problems, nonlinear, underactuated, and non-smooth control tasks, the proposed method demonstrates faster convergence, a higher effective sample size, and greater robustness in low-sample regimes compared to vanilla MPPI and CMA-ES.
These results show that exploiting model structure can improve the efficiency of sampling-based control under tight computational budgets.\\

While our method improves sample efficiency, it introduces some trade-offs.
First, constructing the quadratic model incurs an additional computational cost (see Tab.~\ref{tab:timing}). The magnitude of this overhead depends on the approximation strategy. With the development of fast differentiable physics engines~\cite{DBLP:conf/rss/WerlingOLEL21, brax, DBLP:journals/ral/LidecKLSC21,lidec2025endtoendhighlyefficientdifferentiablesimulation}, obtaining first-order gradients is becoming increasingly negligible, effectively mitigating the cost of analytical or Gauss-Newton approximations. 
In non-differentiable settings that require randomized smoothing, the overhead is distinct.
While these evaluations are fully parallelizable, they incur a per-iteration computational cost that exceeds that of Vanilla MPPI. 
%
% Second, the method relies on the local validity of the quadratic approximation. If the prior mean is far from the basin of attraction, or if the smoothing parameter $\sigma$ is ill-suited to the landscape, the guided prior may converge to local minima, a limitation shared with classical Newton-based methods.
Second, the impact of the quality of the quadratic approximation on performance can be quantified through the residual.
When the model $m^k$ accurately captures the local geometry, the residual $r^k$ has smaller magnitude and lower variance than $f$, the importance weights $\exp(-r^k/\lambda)$ are more uniform, and the effective sample size (ESS, Eq.~\eqref{eqn:ESS}) increases. This is the regime in which our method provides the largest benefit.
Conversely, when the model is  less informative (poor curvature estimates or mismatched smoothing scale), $r^k$ becomes larger and more variable, the weights concentrate, and ESS drops. The sampler then corrects for a poor proposal rather than exploiting a good one, and performance degrades toward vanilla MPPI.
As an illustration, in Fig.~\ref{fig:quadratic-sin}, a too local model would guide the proposal toward the nearest valley, leading to limited benefits or oscillatory behavior. An appropriately coarse model instead smooths out local irregularities and steers iterations toward the correct basin before switching to a finer model for high-precision convergence.

We will focus on reducing computational overheads and expanding the applicability of our model-guided MPPI. We plan to investigate adaptive smoothing strategies in which the noise scale $\sigma$ and temperature $\lambda$ are adapted online based on the residual variance, thereby removing the need for manual heuristic tuning.
Furthermore, we aim to integrate learned models using data-driven priors to guide efficient sampling and to scale this method to high-dimensional, contact-rich domains such as dexterous manipulation, where the cost of exploration is prohibitively high.

% \begin{comment}
\section*{Acknowledgments}
This work has received support from the French government, managed by the National Research Agency, under the France 2030 program with the references Organic Robotics Program (PEPR O2R), “PR[AI]RIE-PSAI” (ANR-23-IACL-0008) and RODEO (ANR-24-CE23-5886).
The European Union also supported this work through the ARTIFACT project (GA no.101165695) and the AGIMUS project (GA no.101070165).
The Paris Île-de-France Région also supported this work in the frame of the DIM AI4IDF.
Views and opinions expressed are those of the author(s) only and do not necessarily reflect those of the funding agencies.
% \end{comment}

%% Use plainnat to work nicely with natbib. 

\bibliographystyle{plainnat}
\bibliography{refs}

\end{document}